\documentclass{article}

\PassOptionsToPackage{numbers, compress}{natbib}

\usepackage[preprint]{neurips_2026}      

\usepackage[utf8]{inputenc}
\usepackage[T1]{fontenc}
\usepackage{hyperref}
\usepackage{url}
\usepackage{booktabs}
\usepackage{amsfonts}
\usepackage{amsmath}
\usepackage{amssymb}
\usepackage{nicefrac}
\usepackage{microtype}
\usepackage{xcolor}
\usepackage{multirow}
\usepackage{listings}
\usepackage{graphicx}
\usepackage{enumitem}
\usepackage{colortbl}
\usepackage{array}
\usepackage{tabularx}
\newcolumntype{Y}{>{\centering\arraybackslash}p{1.4em}}

\title{Graphs of Research: Citation Evolution Graphs\\
as Supervision for Research Idea Generation}

\author{%
  Songyang Gao\thanks{Equal contribution.} \\
  The Hong Kong University of \\ Science and Technology (Guangzhou) \\
  \texttt{sgao068@connect.hkust-gz.edu.cn} \\
  \And
  Yinghui Xia\footnotemark[1] \\
  The Hong Kong University of \\ Science and Technology (Guangzhou) \\
  \texttt{yxia501@connect.hkust-gz.edu.cn} \\
  \And
  Siyi Liu\\
  Tsinghua University \\
  \texttt{liusiyi25@mails.tsinghua.edu.cn} \\
  \And
  Hui Xiong\thanks{Corresponding author.} \\
  The Hong Kong University of \\ Science and Technology \\
  \texttt{xionghui@ust.hk} \\
}

\begin{document}

\maketitle

\begin{abstract}
Research idea generation is the innovation-driving step of automated scientific research. Recently, large language models (LLMs) have shown potential for automating idea generation at scale. However, existing methods mainly condition LLMs on eliciting idea generation through static retrieval of relevant literature or complex prompt engineering, without discarding the structural relations among references. We propose \emph{Graphs of Research} (GoR), a supervised fine-tuning method that extracts a 2-hop reference neighborhood for each seed paper, derives the relations among those references from citation position, frequency, predecessor links, and publication time, and organizes them into a paper-evolution directed acyclic graph (DAG). We construct an automated extraction pipeline that draws data from five major ML/NLP venues, comprising 498/50/50 train/validation/test seed papers and approximately 7{,}600 cited references. Qwen2.5-7B-Instruct-1M is fine-tuned on a structured-text prompt that includes the citation graph, edge signals, reference information, and task definition to predict the idea for the seed paper. Across head-to-head LLM-judge tournaments against gpt-4o-driven baselines, GoR-SFT achieves SOTA, demonstrating the effectiveness of citation-evolution graphs as supervision signal for LLM-based idea generation. We hope that this reduces the barrier for citation evolution graphs as a supervision, accelerating automated scientific innovation.
\end{abstract}

\section{Introduction}
\label{sec:intro}

\begin{figure}[t]
\vspace{-1.0em}
\centering
\includegraphics[width=\linewidth]{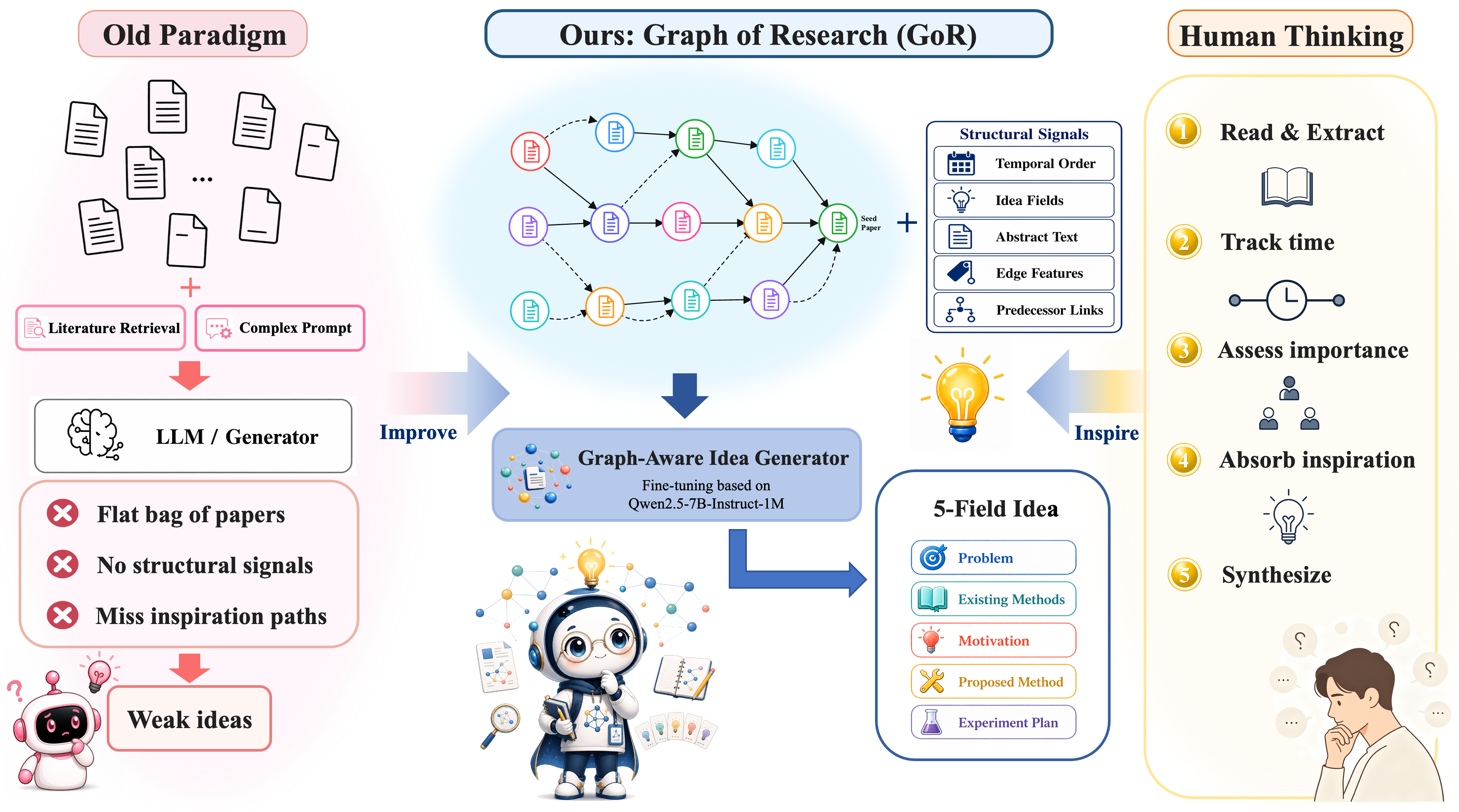}
\caption{Comparison of the existing paradigm of idea generation, our GoR, and the human ideation process inspiring our design.}
\label{fig:motivation}
\vspace{-1.0em}
\end{figure}

Automated scientific research increasingly relies on large language models (LLMs) to compose literature review, ideation, experimental validation, and paper writing into closed loops~\citep{lu2024aiscientist, yamada2025aiscientistv2, tang2025airesearcher, schmidgall2025agentlab}. Within this loop, ideation is the innovation-driving step. It sets the originality and feasibility of every downstream artifact, and remains the one stage where automation cannot fall back on retrieval or rote execution. With LLMs making research ideation tractable at scale~\citep{si2024canllms, li2025chain}, a concrete question follows: \emph{what input lets an LLM generate high-quality, innovative research ideas?}

Existing methods inject information into LLMs to inspire idea generation through static retrieval, agent orchestration, or trained generators. \emph{Retrieval-then-generate} pipelines~\citep{wang2024scimon, hu2024nova, li2025chain} inject neighboring papers as inspiration sources without any training, but rely on retrieval scoring that captures topical similarity. \emph{Multi-agent autonomy} frameworks~\citep{tang2025airesearcher, lu2024aiscientist, yamada2025aiscientistv2, su2025virsci, schmidgall2025agentlab, ghafarollahi2025sciagents, baek2025researchagent} cover the full research lifecycle but reduce ideation itself to repeated prompting and filtering. \emph{Trained policies}~\citep{weng2025cycleresearcher} internalize reviewer preferences through cycle-trained generation, yet supervise at the manuscript level rather than the idea level. As shown in Figure~\ref{fig:motivation} (left), despite their architectural diversity, these systems consume references as a flat text bag, projecting away the structural relations that connect those references in the source literature. In contrast, human researchers read references through structural cues such as section placement, year arithmetic, predecessor relations, and parallel-work patterns, and synthesize these cues into the next idea.

Inspired by this human ideation process, we incorporate these inter-paper structural signals into the supervision pipeline. We propose Graphs of Research (GoR), illustrated in Figure~\ref{fig:motivation} (middle), a structured-text prompt format that serializes each paper's citation subgraph along with edge features and predecessor relations. Concretely, we extract a 2-hop reference neighborhood for each paper, annotate each edge with eight features spanning position provenance, influence, temporal, and structural signals, mark parallel or explicit predecessor relations among nodes, and serialize the annotated graph into a structured-text prompt. We then fine-tune Qwen2.5-7B-Instruct-1M on this prompt with completion-only cross-entropy on the paper's five-field idea, calling the resulting model GoR-SFT. To isolate the effect of structural signals, we train a paired plain-reference baseline (Refs-SFT) on the same 498 training papers from NeurIPS, ICLR, CVPR, ICML, and ACL between 2020 and 2024, with the graph annotations stripped under matched hyperparameters and the structural blocks as the single experimental delta.

We evaluate GoR-SFT on a leak-free test set drawn from 2025 papers at the same five venues using multi-dimensional metrics, including a five-dimension LLM-judge tournament covering novelty, significance, feasibility, clarity, and effectiveness, surface metrics against the gold five-field idea, and a 10-metric human evaluation. To evaluate effectiveness against traditional methods, we compare GoR-SFT with three published gpt-4o-driven idea generation baselines, ranking first on \textbf{31, 40, and 48 of the 50 seeds} respectively (Section~\ref{sec:exp:main}). Through controlled ablation against zero-shot Qwen2.5-7B-Instruct-1M and the matched-capacity Refs-SFT baseline, we isolate SFT as the dominant driver and graph supervision as a focused additional signal that lifts Significance and Clarity (Section~\ref{sec:exp:ablation}). Given the same graph-format prompt, GoR-SFT wins the head-to-head against the much larger gpt-4o consuming the prompt zero-shot, ranking first on $32$ of $50$ seeds, isolating supervision rather than scale as the active ingredient (Section~\ref{sec:exp:capgap}). We further corroborate these automated rankings with a 10-metric blinded human evaluation by 5 NLP and ML PhD raters, where GoR-SFT wins 5 of 10 metrics including \emph{Overall}. \textbf{Our main contributions are summarized as follows:}

\setlength{\leftmargini}{1em}
\begin{itemize}
\item \textbf{Graphs of Research framework.} We characterize citation graph structure as an underused SFT supervision signal in current LLM-based ideation systems, and propose GoR, a method that returns these signals to the supervision pipeline by serializing the citation subgraph as structured text input for LLM fine-tuning (Section~\ref{sec:method}).
\item \textbf{Automated citation-graph extraction pipeline.} We construct an automated pipeline that builds citation subgraphs, edge-feature annotations, and structured five-field idea targets for 498 training papers from NeurIPS, ICLR, CVPR, ICML, and ACL between 2020 and 2024, with a 50-paper in-domain validation set and a leak-free 50-seed test set drawn from 2025 papers (Section~\ref{sec:exp:setup}).
\item \textbf{Experimental validity.} Extensive experimental results demonstrate that GoR-SFT improves the quality of generated ideas, confirming that injecting citation-evolution graph structure into the supervision pipeline is a simple yet effective recipe for steering LLM-based idea generation toward higher-quality, more innovative outcomes (Section~\ref{sec:exp:main}).
\end{itemize}

\begin{figure}[!b]
\vspace{-1.0em}
\centering
\includegraphics[width=\linewidth]{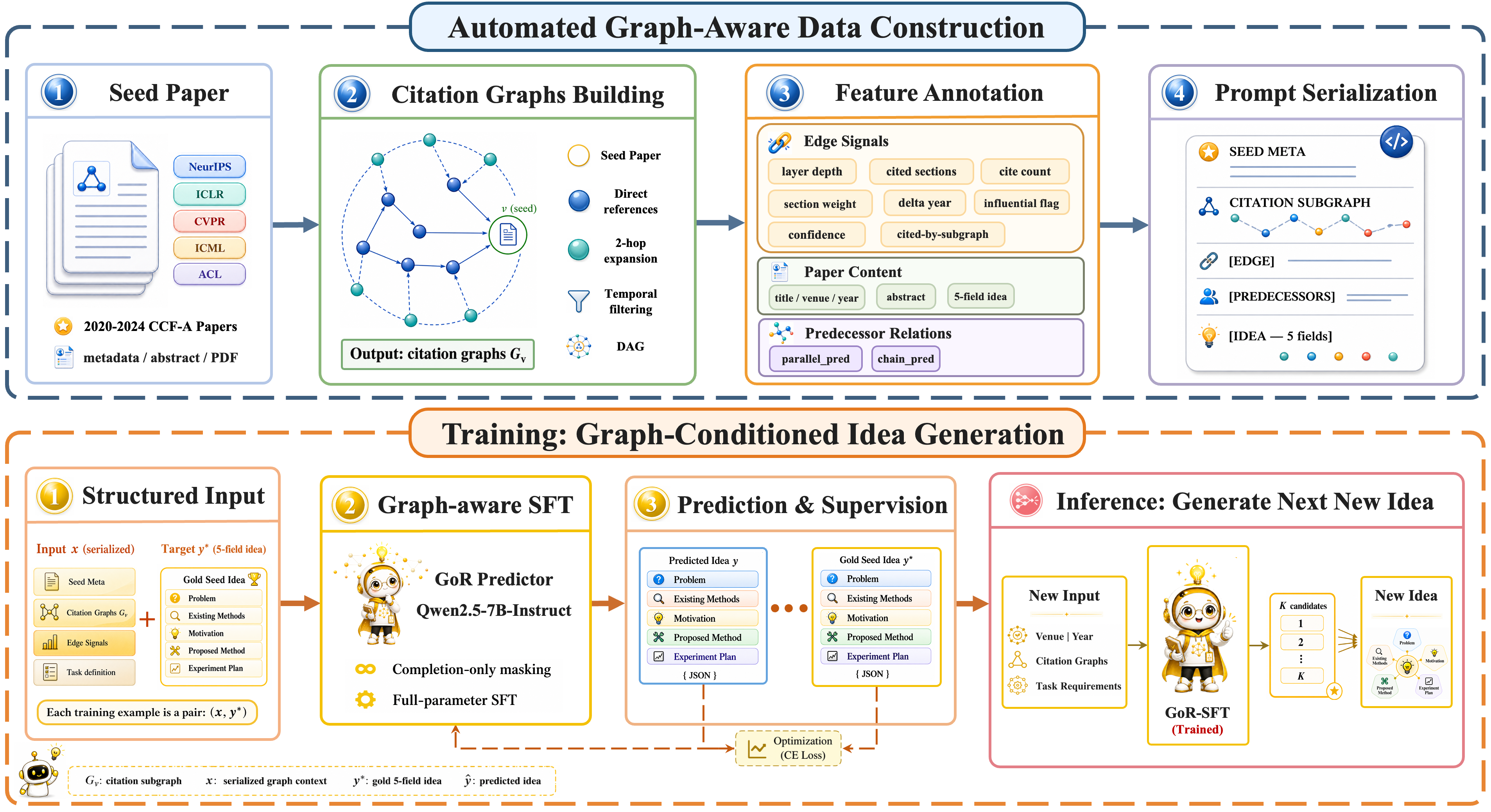}
\caption{\textbf{Our GoR framework.} \emph{Top}: For each seed paper, we extract a citation subgraph, annotate it with eight edge features and parallel or explicit predecessor relations, and serialize the annotated graph into a structured-text prompt (§\ref{sec:method:data}). \emph{Bottom}: We fine-tune Qwen2.5-7B-Instruct-1M on the prompt with completion-only cross-entropy on the seed's five-field idea (§\ref{sec:method:training}), and at inference, GoR-SFT consumes a new citation graph and emits a new idea (§\ref{sec:method:inference}).}
\label{fig:framework}
\vspace{-1.0em}
\end{figure}

\section{Related work}
\label{sec:related}

\paragraph{LLM-based scientific research ideation.}
Scientific idea generation is a core step in automated research, setting the novelty and feasibility ceiling for downstream experimentation and writing. Recent LLM-based systems ground ideation in prior literature in two main ways. Retrieval-then-generate methods use semantic neighbors or citation context as inspiration~\citep{wang2024scimon, si2024canllms}. Agent systems extend ideation into broader research workflows such as experiment design, code generation, review, and manuscript writing~\citep{lu2024aiscientist, yamada2025aiscientistv2, tang2025airesearcher, su2025virsci, schmidgall2025agentlab, ghafarollahi2025sciagents, baek2025researchagent}. More targeted systems improve how prior work is exposed to the generator. ResearchAgent uses academic-graph and entity-level context~\citep{baek2025researchagent}, CoI organizes papers into development chains~\citep{li2025chain}, Nova expands knowledge acquisition through iterative planning and search~\citep{hu2024nova}, and FlowPIE couples literature exploration with test-time idea evolution~\citep{wang2026flowpie}. These methods mainly optimize inference-time retrieval, agent interaction, search, or review. GoR instead asks whether citation-evolution structure in prior literature can be converted into training-time supervision for the generator itself.

\paragraph{Graph-augmented LLMs and citation networks.}
Graph structure has long been used to model scientific knowledge, from Literature-Based Discovery to citation-graph mining, document representation, and impact forecasting~\citep{swanson1986lbd, sybrandt2020agatha, cohan2022specter2, gu2024impact4cast}. Recent graph-augmented LLMs use graphs as retrieval substrates, external memory, or architectural inputs. GraphRAG, LightRAG, and HippoRAG retrieve over graph-structured knowledge~\citep{edge2024graphrag, guo2024lightrag, gutierrez2024hipporag}, while GraphGPT and LLaGA inject graph information through encoders or projector layers~\citep{tang2024graphgpt, chen2024llaga}. Scientific ideation systems also use structured literature relations, including CoI chains, ResearchAgent's academic graph, and FlowPIE's test-time literature graph~\citep{li2025chain, baek2025researchagent, wang2026flowpie}. These works show that flat paper lists discard useful relational signals, but mostly use graphs for retrieval, search, memory, or architecture design. GoR instead serializes a citation-evolution graph as structured text, trains a vanilla LLM on this input, and isolates the contribution of graph structure by stripping only structural labels while preserving the same referenced papers.

\section{Method}
\label{sec:method}

In this paper, we model the human research-ideation process by constructing a temporal-evolution DAG over each paper's references, providing the LLM with structural cues for innovative idea generation. As illustrated in Figure~\ref{fig:framework}, GoR operates in three stages. (i)~We extract a citation subgraph for each seed paper and annotate it with eight edge features and parallel or explicit predecessor relations (§\ref{sec:method:data}). (ii)~We fine-tune Qwen2.5-7B-Instruct-1M on the serialized subgraph with completion-only cross-entropy on the seed's five-field idea, calling the resulting model GoR-SFT (§\ref{sec:method:training}). (iii)~At inference, GoR-SFT consumes a new citation graph and emits a structured five-field idea (§\ref{sec:method:inference}).

\begin{figure}[!b]
\vspace{-1.0em}
\centering
\includegraphics[width=0.8\linewidth]{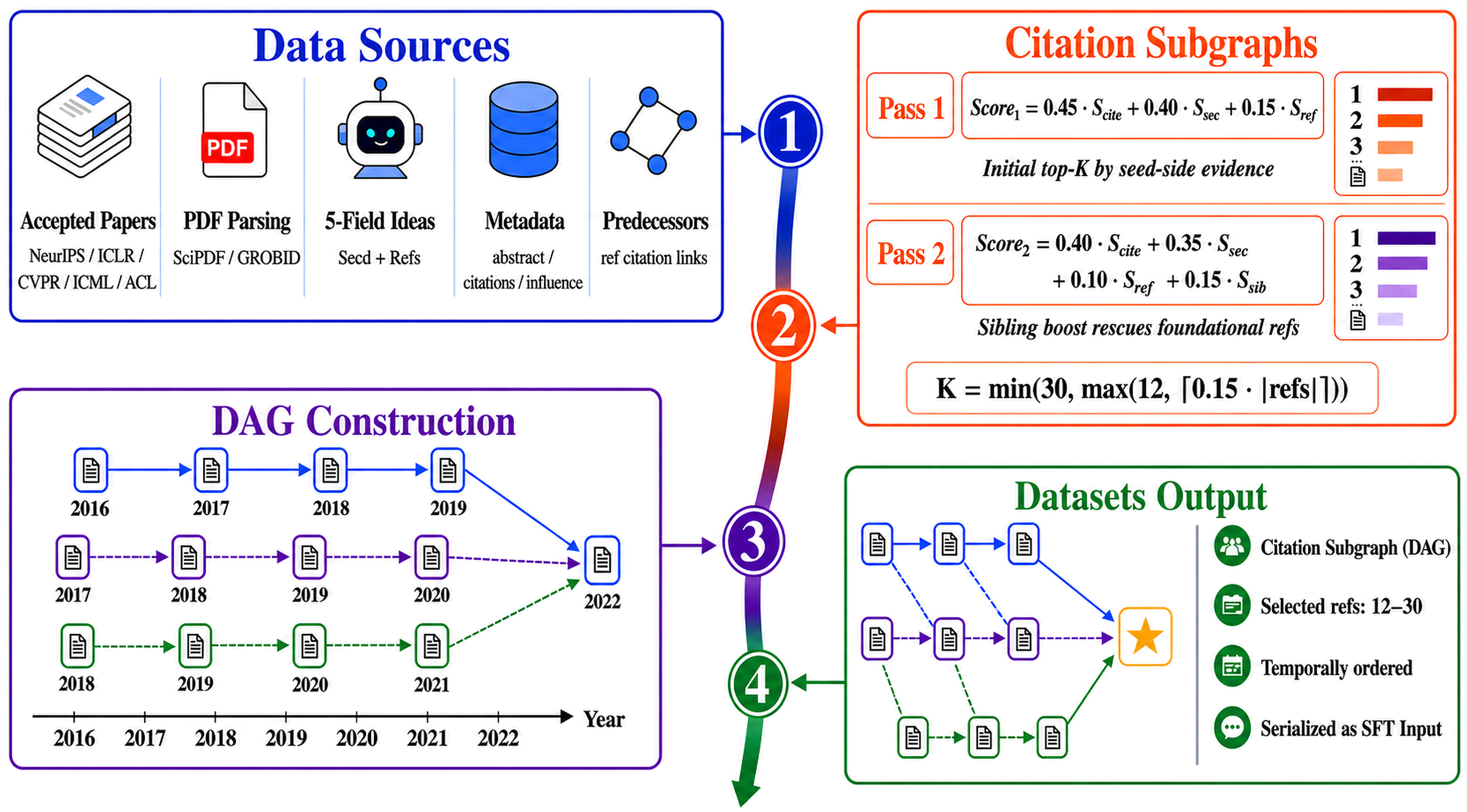}
\caption{\textbf{The pipeline turning a seed paper into an annotated citation DAG.} (1)~\emph{Data sources}: parse PDFs, extract five-field ideas, fetch metadata, mine predecessor edges. (2)~\emph{Citation subgraphs}: rank candidates by seed-side evidence, then re-rank with sibling boost rescuing foundational refs, keeping the top $K \in [12, 30]$. (3)~\emph{DAG construction}: connect surviving refs along the temporal cone backward from the seed, yielding explicit, parallel, and direct-to-seed edges. (4)~\emph{Datasets output}: a temporally ordered subgraph anchored at the seed, ready for serialization as the SFT input.}
\label{fig:subgraph}
\vspace{-1.0em}
\end{figure}

\subsection{Automated graph-aware data construction}
\label{sec:method:data}

To construct the GoR training corpus, we build a citation DAG for each accepted conference paper, with the seed paper as the sink and its references as the other nodes. An in-subgraph scoring policy decides which references are retained, and we annotate every node and edge with structural features. Figure~\ref{fig:subgraph} illustrates the four stages of the pipeline.

\paragraph{Data sources.} For each seed paper we retrieve the PDF (OpenReview when available, arXiv otherwise), parse sections and references with GROBID~\citep{lopez2009grobid}, and fetch reference-side metadata (abstract, year, venue, citation count, \texttt{isInfluential} flag, in-text \texttt{contexts}) from the Semantic Scholar Graph API~\citep{kinney2023s2}. Both the seed and each retrievable reference are extracted by an LLM into a shared five-field idea schema that captures each paper's core content. The five fields are Problem, Existing Methods, Motivation, Proposed Method, and Experiment Plan, with the extractor prompt and a quality audit reported in Appendix~\ref{app:extract}. For each reference, we additionally intersect its own reference list with the seed's reference set to recover in-subgraph predecessor links used downstream by the scoring policy and the DAG.

\paragraph{Citation subgraph.} Let $\mathcal{V}$ denote the universe of papers and $\mathcal{R}(v) \subseteq \mathcal{V}$ the explicit Semantic Scholar references of seed $v$. We form a 2-hop expansion
\begin{equation}
\mathcal{N}(v) \;=\; \mathcal{R}(v) \,\cup\, \bigl\{ r' \,\bigm|\, r' \in \mathcal{R}(r),\, r \in \mathcal{R}(v) \bigr\},
\end{equation}
and enforce a strict temporal cone, retaining
\begin{equation}
G_v \;=\; \bigl\{ u \in \mathcal{N}(v) \,\bigm|\, t(u) < t(v) \bigr\}.
\end{equation}
This construction is identical to a standard time-split LBD setup modulo the per-paper truncation introduced next.

\paragraph{DAG construction.} We prune $G_v$ to a budget $K = \min\!\bigl(30,\, \max(12,\, \lceil 0.15\,|\mathcal{R}(v)|\rceil)\bigr)$, chosen so the median serialized prompt fits inside the 16k-token training context of Qwen2.5-7B-Instruct-1M. Pruning runs in two passes. Pass~1 ranks each candidate $r$ by
\begin{equation}
\mathrm{Score}_1(r) \;=\; 0.45\, S_{\mathrm{cite}}(r) + 0.40\, S_{\mathrm{sec}}(r) + 0.15\, S_{\mathrm{infl}}(r),
\end{equation}
combining seed-internal cite frequency, section weight, and the Semantic Scholar influential flag. Pass~2 adds a sibling-boost term
\[
S_{\mathrm{sib}}(r) \;=\; \bigl|\{\, s \in \mathrm{TopK}_1 : r \in s.\mathrm{predecessors}\,\}\bigr| \,/\, K,
\]
which surfaces canonical ancestors with low local cite count yet referenced by many already-selected papers (e.g.\ GPT-1 within the GPT-3 subgraph). The final rank is
\begin{equation}
\mathrm{Score}_2(r) \;=\; 0.40\, S_{\mathrm{cite}} + 0.35\, S_{\mathrm{sec}} + 0.10\, S_{\mathrm{infl}} + 0.15\, S_{\mathrm{sib}}.
\end{equation}
We then connect every pair $(u, v)$ with $u \in \mathrm{predecessors}(v)$ by an edge classified as \texttt{explicit\_pred} when $t(v) > t(u)$, \texttt{parallel\_pred} when $t(v) = t(u)$, or \texttt{direct\_to\_seed} for sources without an in-subgraph predecessor. 2-cycles are broken by $S_{\mathrm{cite}}$ with paper-id lexicographic order as a tie-breaker. Every retained node carries its title, year, venue, abstract, and five-field idea. Every edge carries eight features grouped into five categories, namely \emph{role} (which seed sections cite $u$), \emph{influence} (Semantic Scholar's flag plus subgraph-local centrality), \emph{recency} (year arithmetic), \emph{topology} (hop distance), and \emph{provenance} (parser-side confidence). The full feature list with sources and semantics appears in Appendix~\ref{app:edge-features}.

\subsection{Graph-conditioned SFT}
\label{sec:method:training}

We train Qwen2.5-7B-Instruct-1M~\citep{yang2024qwen25} on a structured-text serialization of the annotated subgraph $G_v$ from §\ref{sec:method:data} to predict the seed paper's five-field idea, calling the resulting model GoR-SFT.

\paragraph{Prompt serialization.} We serialize $G_v$ into a structured-text prompt with one block per node (\texttt{[EDGE]}, \texttt{[PREDECESSORS]}, \texttt{[IDEA]}, \texttt{[ABSTRACT]}) emitted in temporal order, followed by a TASK block requesting the seed's five-field idea in strict JSON. Full template in Appendix~\ref{app:prompt}.

\paragraph{Training objective.} Let $x = \mathrm{serialize}(G_v)$ be the prompt and $y$ the gold five-field JSON for seed $v$. We optimize standard completion-only cross-entropy
\begin{equation}
\mathcal{L}(\theta) \;=\; -\sum_{(x, y) \in \mathcal{D}} \sum_{t=1}^{|y|} \log p_\theta\!\bigl(y_t \,\bigm|\, x,\, y_{<t}\bigr),
\end{equation}
masking all prompt tokens via \texttt{DataCollatorForCompletionOnlyLM} so gradient flows only on the JSON answer. The single deliberate delta between \texttt{GoR-SFT} and the matched \texttt{Refs-SFT} baseline is the presence of \texttt{[EDGE]} and \texttt{[PREDECESSORS]} blocks in $x$. Both runs share the base model, optimizer, schedule, batch size, precision, and random seed. Headline settings are 2 epochs of full fine-tuning, effective batch size 8, learning rate $2 \times 10^{-5}$ on a cosine-with-min-lr schedule, bfloat16 precision, and \texttt{max\_seq\_len} 16{,}384 on $4\times$A800-80G GPUs. The full hyperparameter table is in Appendix~\ref{app:hp}. We perform NLL-based checkpoint selection on a held-out 50-paper in-domain validation set described in §\ref{sec:exp:setup}.

\subsection{Next idea generation}
\label{sec:method:inference}

At inference, GoR-SFT consumes a new seed's citation subgraph and decodes a five-field idea with vLLM~\citep{kwon2023vllm}, sampling $K{=}10$ candidates per prompt at temperature 0.9 and top-$p$ 0.95. The LLM-judge tournament uses one decoded idea per method per seed, with surface metrics using oracle top-1 selection over the $K$ candidates (§\ref{sec:exp:setup}, results in §\ref{sec:exp:main}).

\section{Experiments}
\label{sec:exp}

\subsection{Research questions}
\label{sec:exp:rq}

We organize our empirical studies of the GoR framework around three research questions on LLM-based idea generation.

\begin{itemize}[leftmargin=1.2em, topsep=2pt, itemsep=2pt]
\item \textbf{RQ1: Input paradigm comparison.} How do different paradigms for presenting prior literature to an LLM affect idea generation quality across the five evaluation dimensions of novelty, significance, feasibility, clarity, and effectiveness?

\item \textbf{RQ2: Component contribution analysis.} Within the GoR pipeline at matched 7B capacity, what are the individual contributions of supervised fine-tuning and the structural graph input to overall idea generation quality?

\item \textbf{RQ3: Supervision versus capacity.} Can a 7B model trained with paper-evolution citation graphs as supervision match or exceed a much larger closed-source LLM that consumes the same graph-format input in zero-shot mode?
\end{itemize}

\subsection{Experimental setup}
\label{sec:exp:setup}

\paragraph{Data.} Following the construction pipeline described in Section~\ref{sec:method:data}, we build the GoR dataset from accepted papers at five major NLP and ML venues: NeurIPS, ICLR, CVPR, ICML, and ACL. The training pool contains 498 papers from 2020 to 2024, with the per-year distribution in Appendix~\ref{app:data-stats}. We hold out a 50-paper in-domain validation set from the same year span and venue mix for NLL-based checkpoint selection. The test set comprises 50 in-distribution seeds drawn from accepted 2025 papers at the same five venues, distributed as ICML (13), ACL (12), NeurIPS (9), ICLR (9), and CVPR (7), with a median of 64 references per seed (range 24--151). The strict temporal cone $t(\text{test}) > t(\text{train})$ ensures no test seed appears as a node or 2-hop reference in any training subgraph. We additionally verified leak-free via title-string and Semantic-Scholar paper-id overlap checks.

\paragraph{Baselines.} We compare our two GoR variants against three published works on LLM-based research idea generation. The two variants differ in how they consume the citation graph. \texttt{GoR-SFT} fine-tunes a 7B Qwen base on the graph-format prompt, while \texttt{GoR-Agent} feeds the same graph-format prompt to gpt-4o at inference without supervision. To ensure a fair comparison, all three baselines use gpt-4o as the shared LLM backbone, run on the same 50-seed test set, and produce the same five-field idea schema, which minimizes judge preference toward more structured outputs. We compare against the following baselines:
\begin{itemize}[leftmargin=1.2em, topsep=2pt, itemsep=2pt]
\item \textbf{Si baseline}~\citep{si2024canllms}: retrieval-augmented over-generate-and-rerank with pairwise comparison.
\item \textbf{CoI-Agent}~\citep{li2025chain}: chronological chain-of-ideas and generates new ideas grounded in this evolution.
\item \textbf{ResearchAgent}~\citep{baek2025researchagent}: multi-agent reviewer-critique loop with academic-graph entity grounding.
\end{itemize}

\paragraph{Models.} Three LLMs serve different roles: Qwen2.5-7B-Instruct-1M~\citep{yang2024qwen25} as our base to train \texttt{GoR-SFT}, gpt-4o as the shared backbone for the three baselines and our \texttt{GoR-Agent} variant, and DeepSeek-V3.2-Exp~\citep{deepseekv32} as the fixed LLM judge.

\paragraph{Evaluation protocols.} To validate GoR, we conduct two automated evaluations (T1 and T2) and one human evaluation (H). \textbf{T1: LLM-judge tournament.} For each test seed, the LLM judge scores all ordered pairs in an $N$-way round-robin along the five dimensions of novelty, significance, feasibility, clarity, and effectiveness. Each pair receives 2/1/0 per dimension following the protocol of \citet{li2025chain}, with both presentation orderings. We report mean Elo, per-method rank-1 count, and per-dimension winrate. \textbf{T2: surface metrics.} For each test seed, we sample $K{=}10$ candidates at temperature 0.9 and top-$p$ 0.95, then report three surface metrics against the gold five-field idea: SPECTER2~\citep{cohan2022specter2} weighted top-1 cosine (wTop1), Method-field ROUGE-L F1~\citep{lin2004rouge} (mROUGE), and DeBERTa-based BERTScore F1~\citep{he2021deberta, zhang2020bertscore} (BERT-F1), all higher-is-better. \textbf{H: human evaluation.} Three NLP and ML PhD raters score the four systems on 10 metrics on a 1--10 Likert scale across a balanced 5-seed blinded subset. The first five dimensions align with T1, and the remaining five (Excitement, Soundness, Originality, Reproducibility, Overall) are drawn from prior human studies~\citep{baek2025researchagent, weng2025cycleresearcher}. Full protocol and per-dimension results appear in Appendix~\ref{app:human-protocol}.

\begin{figure}[!b]
\vspace{-1.0em}
\centering
\includegraphics[width=0.75\linewidth]{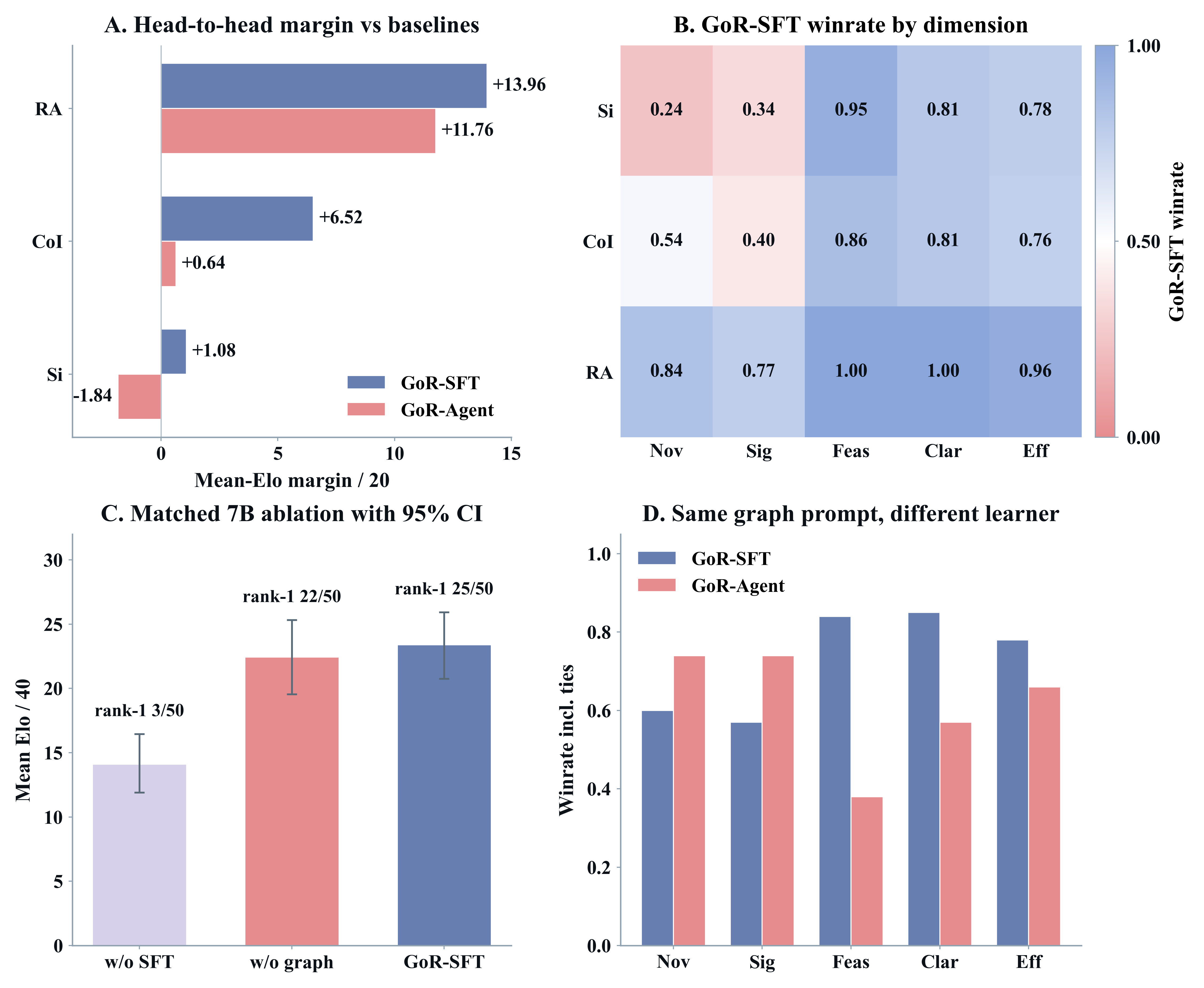}
\caption{(A) Mean Elo margins of \texttt{GoR-SFT} and \texttt{GoR-Agent} against the three published baselines on the test set. (B) Per-dimension winrates of \texttt{GoR-SFT} against the three baselines. (C) Mean Elo of the three matched 7B variants \texttt{w/o SFT}, \texttt{w/o graph}, and \texttt{GoR-SFT}, with 95\% bootstrap CIs over seeds. (D) Per-dimension winrates of \texttt{GoR-SFT} versus \texttt{GoR-Agent} on the same graph input.}
\label{fig:exp-summary}
\vspace{-1.0em}
\end{figure}

\subsection{Main results: head-to-head against published baseline agents}
\label{sec:exp:main}

To answer RQ1, GoR-SFT and GoR-Agent each compete against three gpt-4o-driven baselines in pairwise LLM-judge tournaments. GoR-Agent isolates the input paradigm under the shared gpt-4o backbone, while GoR-SFT tests whether SFT on the same graph prompt recovers further gains.

\begin{table}[!ht]
\caption{\textbf{Main results: GoR-Agent and GoR-SFT vs published baselines on the test set.} Bold marks the GoR variant's wins.}
\label{tab:t1-main}
\centering
\setlength{\tabcolsep}{3pt}
\renewcommand{\arraystretch}{1.1}
\resizebox{\linewidth}{!}{%
\begin{tabular}{l*{15}{Y}cc}
\toprule
& \multicolumn{3}{c}{\textbf{Nov}} & \multicolumn{3}{c}{\textbf{Sig}} & \multicolumn{3}{c}{\textbf{Feas}} & \multicolumn{3}{c}{\textbf{Clar}} & \multicolumn{3}{c}{\textbf{Eff}} & & \\
\cmidrule(lr){2-4} \cmidrule(lr){5-7} \cmidrule(lr){8-10} \cmidrule(lr){11-13} \cmidrule(lr){14-16}
\textbf{Opponent} & W & T & L & W & T & L & W & T & L & W & T & L & W & T & L & \textbf{Rank-1} & \textbf{Mean Elo}/20 \\
\midrule
\rowcolor{cyan!10}
\multicolumn{18}{c}{\textit{\texttt{GoR-Agent}: gpt-4o backbone with graph-format prompt (zero-shot)}} \\
Si baseline             & 0.00           & 0.26 & 0.74 & 0.06           & 0.30 & 0.64 & \textbf{0.76} & 0.12 & 0.12 & \textbf{0.38} & 0.34 & 0.28 & \textbf{0.34} & 0.36 & 0.30 & \textbf{26} / 24 & 9.08 / 10.92            \\
CoI-Agent               & 0.16          & 0.42 & 0.42 & 0.20          & 0.32 & 0.48 & \textbf{0.48} & 0.22 & 0.30 & \textbf{0.54} & 0.24 & 0.22 & \textbf{0.48} & 0.20 & 0.32 & \textbf{31} / 19 & \textbf{10.32} / 9.68   \\
ResearchAgent           & \textbf{0.38} & 0.46 & 0.16 & \textbf{0.44} & 0.28 & 0.28 & \textbf{0.92} & 0.04 & 0.04 & \textbf{0.96} & 0.02 & 0.02 & \textbf{0.84} & 0.10 & 0.06 & \textbf{47} / 3  & \textbf{15.88} / 4.12   \\
\rowcolor{cyan!10}
\multicolumn{18}{c}{\textit{\texttt{GoR-SFT}: 7B base fine-tuned on the same graph supervision}} \\
Si baseline             & 0.04           & 0.20 & 0.76 & 0.12          & 0.24 & 0.64 & \textbf{0.88} & 0.10 & 0.02 & \textbf{0.56} & 0.28 & 0.16 & \textbf{0.48} & 0.38 & 0.14 & \textbf{31} / 19 & \textbf{10.54} / 9.46   \\
CoI-Agent               & 0.32          & 0.24 & 0.44 & 0.24          & 0.28 & 0.48 & \textbf{0.78} & 0.16 & 0.06 & \textbf{0.76} & 0.16 & 0.08 & \textbf{0.70} & 0.16 & 0.14 & \textbf{40} / 10 & \textbf{13.26} / 6.74   \\
ResearchAgent           & \textbf{0.58} & 0.28 & 0.14 & \textbf{0.54} & 0.26 & 0.20 & \textbf{0.94} & 0.06 & 0.00 & \textbf{0.98} & 0.02 & 0.00 & \textbf{0.86} & 0.12 & 0.02 & \textbf{48} / 2  & \textbf{16.98} / 3.02   \\
\bottomrule
\end{tabular}%
}
\end{table}

\paragraph{Graph as a zero-shot input alone is unstable.} GoR-Agent wins two of the three tournaments on mean Elo, with 10.32 against CoI-Agent and 15.88 against ResearchAgent, but loses to Si baseline at 9.08 vs 10.92 in the top three rows of Table~\ref{tab:t1-main}, with the same margins visualized in Figure~\ref{fig:exp-summary}(A). The pattern is sharper at the dimension level. GoR-Agent wins decisively on the three structural dimensions Feasibility, Clarity, and Effectiveness, where its per-seed win rate spans 0.34 to 0.96 across the three opponents. The asymmetry traces to prompt paradigm: Si baseline's prompt explicitly encourages open-ended divergent ideation over a flat reference bank, while the GoR-Agent prompt constrains the model to derive ideas from the structured citation evolution graph. The structural constraint thus trades raw novelty for grounded feasibility. This pattern raises the question of whether prompt-level injection suffices for the model to internalize the graph signal, or whether supervised fine-tuning on the graph is required. We address this with GoR-SFT next.

\paragraph{SFT on graph supervision yields consistent SOTA.} When the graph signal is learned at training time, GoR-SFT wins all three head-to-head tournaments on mean Elo, with 10.54 against Si baseline, 13.26 against CoI-Agent, and 16.98 against ResearchAgent, ranking first on \textbf{31, 40, and 48 of the 50 test seeds} respectively (bottom three rows of Table~\ref{tab:t1-main}, Figure~\ref{fig:exp-summary}(A)). On Feasibility, Clarity, and Effectiveness, per-seed win rate spans 0.48 to 0.98 (Figure~\ref{fig:exp-summary}(B)). On creative dimensions GoR-SFT outperforms ResearchAgent and CoI-Agent on Novelty (0.58, 0.32) and Significance (0.54), while losing only to Si baseline on Novelty (0.04) and Significance (0.12).

\subsection{Ablation: structural input is the driver}
\label{sec:exp:ablation}

To answer RQ2, we decompose \texttt{GoR-SFT}'s gain under matched 7B capacity: how much comes from SFT over the zero-shot base, and how much comes from the structural input on top of plain-reference SFT? Table~\ref{tab:t2-main} merges the 3-way LLM-judge tournament across three 7B-Instruct ablation variants (\texttt{w/o SFT}, \texttt{w/o graph}, \texttt{GoR-SFT}) with T2 surface metrics on the same 50-seed test set.

GoR-SFT achieves mean Elo 23.42 out of 40 in the 3-way tournament, exceeding plain-reference \texttt{w/o graph} at 22.46 and zero-shot \texttt{w/o SFT} at 14.12, and ranks first on 25 of the 50 test seeds compared to 22 for \texttt{w/o graph} and 3 for \texttt{w/o SFT}. The pattern is sharper at the dimension level. GoR-SFT wins 4 of 5 dimensions over \texttt{w/o graph}, with Significance reaching 0.730 against 0.625 and Clarity 0.775 against 0.705, while Feasibility is essentially tied at 0.640 against 0.645. The same ordering holds on the T2 surface metrics, where GoR-SFT wins all three of wTop1, mROUGE, and BERT-F1. The T2 margins are smaller as wTop1 saturates near 0.93 across SFT variants, while the LLM-judge tournament discriminates more sharply on idea content.

\begin{table}[!ht]
\caption{\textbf{Quantitative comparison of \texttt{w/o SFT}, \texttt{w/o graph}, and \texttt{GoR-SFT}.} Best per column in \textbf{bold}.}
\label{tab:t2-main}
\centering
\scriptsize
\setlength{\tabcolsep}{3pt}
\begin{tabular}{lcccccrccccc}
\toprule
\textbf{Method} & \textbf{Nov} & \textbf{Sig} & \textbf{Feas} & \textbf{Clar} & \textbf{Eff} & \textbf{Rank-1} & \textbf{Mean Elo}/40 & \textbf{wTop1}\,$\uparrow$ & \textbf{mROUGE}\,$\uparrow$ & \textbf{BERT-F1}\,$\uparrow$ \\
\midrule
\texttt{w/o SFT}             & 0.480 & 0.470 & 0.520 & 0.505 & 0.480 & 3/50 & 14.12 & 0.9256 & 0.1187 & 0.5871 \\
\texttt{w/o graph}           & 0.715 & 0.625 & \textbf{0.645} & 0.705 & 0.735 & 22/50 & 22.46 & 0.9280 & 0.1386 & 0.5990 \\
\texttt{GoR-SFT}\,($\star$)  & \textbf{0.740} & \textbf{0.730} & 0.640 & \textbf{0.775} & \textbf{0.745} & \textbf{25/50} & \textbf{23.42} & \textbf{0.9290} & \textbf{0.1393} & \textbf{0.6026} \\
\bottomrule
\end{tabular}
\end{table}

Figure~\ref{fig:exp-summary}(C) summarizes the two contributions in mean Elo terms. The SFT contribution is large and statistically stable. GoR-SFT reaches mean Elo 23.42 against zero-shot Qwen's 14.12, a margin of 9.30 with paired bootstrap 95\% CI [5.32, 13.14] and one-sided Wilcoxon $p < 10^{-4}$. The graph contribution on top of plain-reference SFT is smaller. GoR-SFT reaches 23.42 against \texttt{w/o graph}'s 22.46, a margin of 0.96 that we interpret as evidence for a useful structural prior. At the dimension level the two effects have different signatures. The SFT-over-zero-shot effect lifts all five dimensions broadly, with per-dimension gains ranging from 0.125 on Feasibility to 0.255 on Effectiveness. The graph-over-plain effect concentrates on Significance and Clarity at gains of 0.105 and 0.070, with Feasibility essentially unchanged.

\subsection{Supervision versus capacity}
\label{sec:exp:capgap}

To answer RQ3, we conduct a head-to-head LLM-judge evaluation of \texttt{GoR-SFT} against \texttt{GoR-Agent} on the test set. Both systems consume the identical graph-format prompt at inference, isolating the contrast between using the citation graph as a prompt-only injection versus as a training-time supervision signal for idea generation. GoR-SFT wins the pair on mean Elo, 11.10 against 8.90, and ranks first on 32 of the 50 test seeds. The win is concentrated on the structural dimensions Feasibility, Clarity, and Effectiveness at 0.840, 0.850, and 0.780, while gpt-4o's larger parametric memory retains Novelty and Significance at 0.740 each, as reported in Table~\ref{tab:capgap} and visualized in Figure~\ref{fig:exp-summary}(D). GoR-SFT additionally offers a Pareto improvement in the cost-quality tradeoff. The 7B model runs locally and completes the 50-seed test set in 2 to 3 minutes with no API spend, while GoR-Agent and the three gpt-4o-driven baselines all rely on the OpenAI API at roughly \$0.004 per idea.

\begin{table}[!ht]
\caption{\textbf{Quantitative comparison of \texttt{GoR-SFT} and \texttt{GoR-Agent} on the test set under matched graph-format input.} Best per column in bold. Cost is the per-idea inference cost in USD.}
\label{tab:capgap}
\centering
\small
\setlength{\tabcolsep}{4pt}
\begin{tabular}{lccccccrc}
\toprule
\textbf{Method} & \textbf{Nov} & \textbf{Sig} & \textbf{Feas} & \textbf{Clar} & \textbf{Eff} & \textbf{Rank-1} & \textbf{Mean Elo}/20 & \textbf{Cost} \\
\midrule
\texttt{GoR-SFT}\,($\star$)         & 0.600 & 0.570 & \textbf{0.840} & \textbf{0.850} & \textbf{0.780} & \textbf{32/50} & \textbf{11.10} & \textbf{\$0} \\
\texttt{GoR-Agent}                  & \textbf{0.740} & \textbf{0.740} & 0.380 & 0.570 & 0.660 & 18/50 & 8.90 & \$0.004 \\
\bottomrule
\end{tabular}
\end{table}

\subsection{Case study: structure-aware idea generation on a citation subgraph}
\label{sec:exp:case}

\begin{figure}[!b]
\vspace{-0.1em}
\centering
\includegraphics[width=0.9\linewidth]{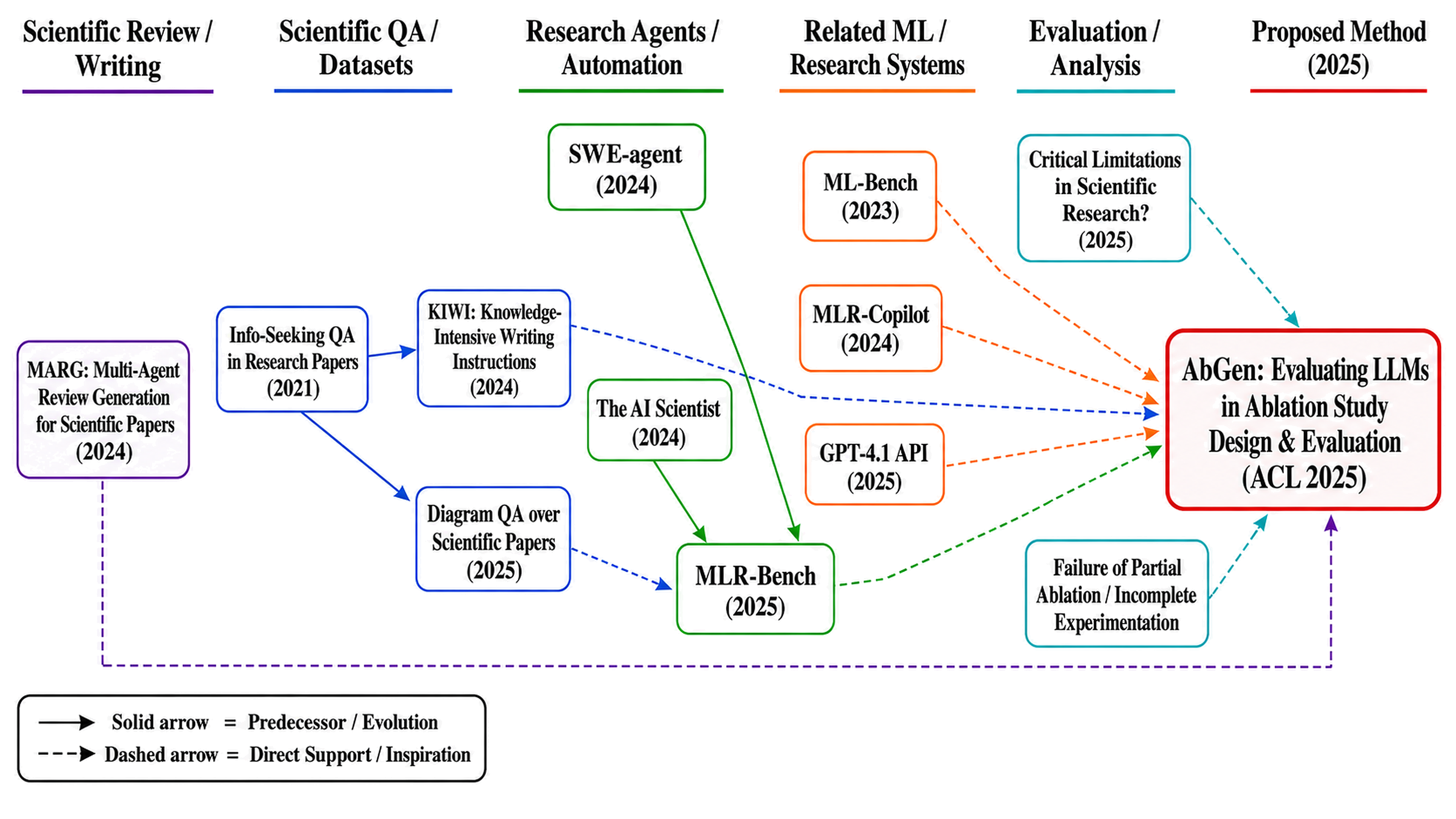}
\caption{\textbf{Citation subgraph for the \emph{AbGen}.} }
\label{fig:case-subgraph}
\vspace{-0.1em}
\end{figure}

We close with a qualitative case study on the seed paper \emph{AbGen}~\citep{zhao2025abgen}. The retained subgraph (Fig.~\ref{fig:case-subgraph}) spans 12 refs across multi-agent review, agent benchmarks, end-to-end agents, and multimodal scientific QA, leaving a gap at the intersection of iterative ablation refinement and long-range research planning that no retained ref addresses jointly.

Reading the four proposals (Table~\ref{tab:case-diff} in Appendix~\ref{app:case-study}), \texttt{GoR-SFT} identifies the refinement-plus-planning gap and proposes a two-component framework anchored in MARG plus a planning module for long-range decisions. Si baseline proposes AGCLO, a free-floating three-stage multi-agent system unanchored to any subgraph ref. CoI-Agent proposes AbGen-XAI+, an extension of AbGen and ML-Bench from the chronological chain. ResearchAgent pivots to FDMME, a multimodal embedding pipeline grounded in the multimodal subset of the subgraph but largely orthogonal to the core ablation-design problem. The LLM-judge verdict reflects the grounding pattern: \texttt{GoR-SFT} wins \textbf{15-5} against Si baseline and \textbf{15-5} against CoI-Agent, and ties \textbf{10-10} against ResearchAgent.

Unlike cases where one paradigm sweeps all baselines, this seed spreads the verdict: \texttt{GoR-SFT} decisively beats the two baselines whose paradigm gives no clean grounding, while it ties ResearchAgent whose entity-store paradigm coincidentally surfaces the multimodal subset of the subgraph. The case nevertheless supports the proposed mechanism. When a coherent structural signal exists in the subgraph, a fine-tuned model identifies it and produces a grounded proposal that two of the three gpt-4o-driven baseline paradigms cannot reproduce.

\section{Conclusion}
\label{sec:conclusion}

We introduce GoR, a supervised fine-tuning recipe that supervises a 7B LLM on citation evolution graphs, built by an automated graph-aware data-construction pipeline, rather than on a flat reference bag. GoR-SFT wins all three head-to-head tournaments against three published baselines, ranking first on 31, 40, and 48 of the 50 seeds respectively. Ablation isolates SFT as the dominant driver, with graph supervision contributing focused additional gains on Significance and Clarity over plain-reference SFT. GoR-SFT further takes 32 of 50 seeds head-to-head against a much larger gpt-4o consuming the same graph, showing that a 7B open-source base with graph supervision can surpass flagship LLMs on idea generation at near-zero inference cost. In a blinded 5-rater human study, GoR-SFT additionally wins 5 of 10 dimensions, independently corroborating the LLM-judge ranking. Citation-graph structure is therefore a useful supervision signal that current LLM-based ideation systems systematically overlook. Looking forward, we plan to scale the automated extraction pipeline to build larger paper-evolution graph datasets, equipping LLMs with a deeper grasp of scientific evolution and accelerating research innovation. We will release the data, training, and evaluation pipeline to encourage extensions to other base LLMs and to hybrid creative-plus-graph prompts.


\bibliographystyle{plainnat}
\bibliography{refs}

\newpage
\appendix

\section*{\centering\Large Appendix}
\addcontentsline{toc}{section}{Appendix}


\section{Five-field idea extraction pipeline}
\label{app:extract}

We extract a structured five-field idea (Problem, Existing Methods, Motivation, Proposed Method, Experiment Plan) from each paper using a prompted-LLM pipeline operating on the paper's introduction and method sections, parsed via GROBID and OpenReview HTML. The extractor follows the schema convention of \citet{baek2025researchagent} and \citet{li2025chain} with a strict-JSON output constraint. The same five-field schema is used uniformly for the 498 training papers, the 50-paper in-domain validation set, and the 50-seed test set, so every paper in our pool has a comparable target representation regardless of its role.

\section{Citation subgraph edge features}
\label{app:edge-features}

Table~\ref{tab:edge-features} lists the eight edge features used by GoR. Every directed edge $u \to v$ in the citation DAG $G_v$ carries these features as labeled key-value pairs, which are emitted in the \texttt{[EDGE]} block of the serialized prompt (Appendix~\ref{app:prompt}). The features cover five categories. \emph{Role} (\texttt{cited\_in\_sections}, \texttt{section\_weight}) describes which seed sections cite the predecessor and how heavily they weight it. \emph{Influence} (\texttt{cite\_count}, \texttt{is\_influential\_raw}, \texttt{cited\_by\_subgraph}) combines a global citation signal from Semantic Scholar with subgraph-local centrality. \emph{Recency} (\texttt{delta\_year}) is plain year arithmetic between seed and predecessor. \emph{Topology} (\texttt{layer\_depth}) records hop distance in the 2-hop neighborhood, distinguishing direct references from references-of-references. \emph{Provenance} (\texttt{low\_confidence}) flags edges whose section attribution comes from a noisy parse and whose value should be discounted by a careful reader. The \emph{w/o graph} ablation strips all eight features along with the \texttt{[PREDECESSORS]} block, leaving only paper title, year, venue, abstract, and the five-field idea.

\begin{table}[h]
\caption{\textbf{The eight edge features in $G_v$ used by GoR.} Source indicates where each feature is derived. Semantics give the intended meaning a graph-aware reader should assign. The \emph{w/o graph} ablation strips all eight features and the predecessor block, leaving only paper title, year, venue, abstract, and the five-field idea.}
\label{tab:edge-features}
\centering
\small
\begin{tabular}{lll}
\toprule
\textbf{Feature} & \textbf{Source} & \textbf{Semantics} \\
\midrule
\texttt{layer\_depth} & hop count $v \to u$ & 1 = direct ref, 2 = ref-of-ref \\
\texttt{cited\_in\_sections} & GROBID / OpenReview parse & seed sections that cite $u$ \\
\texttt{cite\_count} & Semantic Scholar & global citation count of $u$ \\
\texttt{section\_weight} & weighted formula & section-importance score \\
\texttt{delta\_year} & year arithmetic & seed\_year $-$ $u$\_year \\
\texttt{is\_influential\_raw} & Semantic Scholar & seed flags $u$ as influential \\
\texttt{low\_confidence} & parser quality flag & section attribution uncertain \\
\texttt{cited\_by\_subgraph} & $G_v$ self-relation & local centrality of $u$ in $G_v$ \\
\bottomrule
\end{tabular}
\end{table}

\section{Training data year distribution}
\label{app:data-stats}

The 498-paper training pool spans accepted papers at NeurIPS, ICLR, CVPR, ICML, and ACL between 2020 and 2024, with per-year counts of 114 (2020), 102 (2021), 98 (2022), 60 (2023), and 124 (2024). The 50-paper in-domain validation set draws from the same year span and venue mix and is held out only for NLL-based checkpoint selection. The 50-seed test set is drawn from accepted 2025 papers at the same five venues, distributed as ICML (13), ACL (12), NeurIPS (9), ICLR (9), and CVPR (7), and is verified leak-free against the training pool by both title-string and Semantic-Scholar paper-id overlap.

\paragraph{Subgraph-level statistics.} The pipeline yields 498 fully annotated subgraphs. After temporal-cone filtering the median subgraph keeps 12 references (mean 12.7, max 30) and 30 edges (mean 32.8, max 111), distributed on average as 22.6 explicit, 5.6 parallel, and 4.6 direct-to-seed edges. As a running example used in the paper, the GPT-3 paper's 146 references shrink to 22 after Pass~2 scoring and to 18 after temporal-cone filtering, producing 91 edges (55 explicit, 31 parallel, 5 direct-to-seed) and a 7{,}265-token serialized prompt (Appendix~\ref{app:prompt}).

\section{Prompt serialization format}
\label{app:prompt}

Each prompt has the schematic structure shown below, with tokens in angle brackets filled per paper.
\begin{lstlisting}[basicstyle=\ttfamily\scriptsize, frame=single, breaklines=true]
# SEED META
venue: <venue>    year: <year>

# CITATION SUBGRAPH (<n> refs, temporally ordered by year)
## [1] <Title> (<year>, <venue>) authors: <authors>
   [EDGE]
     layer_depth = <int>
     cited_in_sections = <list>
     cite_count = <int>
     section_weight = <float>
     delta_year = <int>
     is_influential_raw = <bool>
     low_confidence = <bool>
     cited_by_subgraph = <int>
   [PREDECESSORS]
     - ref_idx=<list>  delta_yr=<int>  edge_type=<parallel_pred|explicit_pred>
   [IDEA -- 5 fields]
     Problem: ...
     Existing Methods: ...
     Motivation: ...
     Proposed Method: ...
     Experiment Plan: ...
   [ABSTRACT] ...
## [2] ...
...

# TASK
Given the SEED META and CITATION SUBGRAPH above, predict the SEED paper's
five-field idea.

# OUTPUT FORMAT (strict)
{
  "Problem": "...",
  "Existing Methods": "...",
  "Motivation": "...",
  "Proposed Method": "...",
  "Experiment Plan": "..."
}
\end{lstlisting}
The \emph{w/o graph} ablation strips \texttt{[EDGE]} and \texttt{[PREDECESSORS]} blocks and renames \texttt{\#~CITATION SUBGRAPH} to \texttt{\#~REFERENCES}. Every other field is preserved, so the only deliberate delta between the \texttt{GoR-SFT} input and the matched \texttt{Refs-SFT} input is the structural annotation.

\section{Training hyperparameters}
\label{app:hp}

Table~\ref{tab:hp} lists the hyperparameters shared by \texttt{GoR-SFT} and the matched \texttt{Refs-SFT} ablation. The two runs share the base model, optimizer, schedule, batch size, precision, fused-kernel set, and random seed. The single deliberate difference is the presence of the \texttt{[EDGE]} and \texttt{[PREDECESSORS]} blocks in the input prompt, isolating the contribution of structural annotation under matched 7B capacity. We use $4 \times $ A800-80G GPUs in DeepSpeed ZeRO Stage~3 with full fine-tuning, and we report the NLL-best checkpoint on the held-out 50-paper in-domain validation set as the model used for downstream evaluation.

\begin{table}[h]
\caption{\textbf{Shared hyperparameters for both \texttt{GoR-SFT} and \texttt{Refs-SFT}.} The only deliberate difference between the two is the presence of structural blocks in the prompt.}
\label{tab:hp}
\centering
\small
\begin{tabular}{ll}
\toprule
\textbf{Setting} & \textbf{Value} \\
\midrule
Base model & Qwen2.5-7B-Instruct-1M~\citep{yang2024qwen25} \\
Strategy & Full fine-tuning \\
Epochs & 2 \\
Effective batch size & 8 (4 GPU $\times$ device-batch 1 $\times$ grad-accum 2) \\
Learning rate & $2 \times 10^{-5}$, cosine\_with\_min\_lr ($\text{min\_lr\_rate}{=}0.1$), warmup ratio 0.03 \\
Weight decay & 0 \\
Precision & bfloat16 \\
\texttt{max\_seq\_len} & 16{,}384 \\
Gradient checkpointing & enabled \\
Fused kernels & Liger~\citep{liger2024} (RoPE, SwiGLU, RMSNorm), FLCE during training, FlashAttention-2 \\
Random seed & 42 \\
\bottomrule
\end{tabular}
\end{table}

\section{Human evaluation: protocol and results}
\label{app:human-protocol}

We complement the LLM-judge tournament with a small-scale blinded human study comparing four systems (\texttt{GoR-SFT}, Si baseline, CoI-Agent, ResearchAgent) on a balanced 5-seed subset of the 50-seed test set. The 5 seeds were drawn to span widely-cited 2024-2025 papers across the five training venues with varied head-to-head verdicts (3 GoR-favorable, 2 Si-favorable, by mean Elo on T1) and include MLE-bench~\citep{chan2024mlebench}, KAN~\citep{liu2024kan}, rStar-Math~\citep{guan2025rstarmath}, BigCodeBench~\citep{zhuo2024bigcodebench}, and SWE-Lancer~\citep{miserendino2025swelancer}.

\paragraph{Recruitment and anonymization.} We recruited three PhD-track NLP and ML graduate students from the same research community as the authors. None are authors of this paper. Each rater received a private packet containing the 5 topic descriptions (seed paper title and abstract) and the four anonymized 5-field ideas per topic, yielding 600 idea-level scores in total. System labels (A, B, C, D) were independently shuffled per (rater, topic) pair to prevent label leakage across raters. Raters were instructed not to look up the actual seed papers during scoring.

\paragraph{Scoring and metrics.} Raters scored each (topic, system) pair on 10 metrics on a 1--10 integer Likert scale, with anchor descriptions for each metric (1--3, 4--6, 7--8, 9--10). The first five (Novelty, Significance, Feasibility, Clarity, Effectiveness) align with the T1 LLM-judge dimensions to enable cross-population comparison. The second five (Excitement, Soundness, Originality, Reproducibility, Overall) cover reviewer-side properties drawn from \citet{baek2025researchagent} and \citet{weng2025cycleresearcher} and the NeurIPS reviewer guidelines.

\paragraph{Workload and compensation.} Each rater submitted $5 \times 4 \times 10 = 200$ scores plus optional comments. Self-reported task time averaged 3 to 4 hours per rater. The study was an internal academic exercise within the authors' research group, so no IRB protocol was required at our institution and no monetary compensation was offered.

\paragraph{Aggregation.} For each (system, dimension) cell of Table~\ref{tab:human-eval}, we report the mean over $n{=}25$ scores ($5$ seeds $\times$ $5$ raters). For inter-rater agreement we report Krippendorff's $\alpha$ on interval data per dimension, computed by treating each (seed, system) as an item and the five raters as observers. The packet generation script, shuffling seed, and aggregation script are released alongside the paper to enable replication.

\paragraph{Results.} Table~\ref{tab:human-eval} reports the per-dimension means. The 10 metrics partition cleanly along an execution-versus-creativity axis. \texttt{GoR-SFT} wins the four execution-oriented dimensions Feasibility, Clarity, Soundness, and Reproducibility at 6.96, 7.48, 7.00, and 6.92, plus \emph{Overall} at 6.56 against Si baseline's 6.48, CoI-Agent's 6.00, and ResearchAgent's 4.84. The five creativity-oriented dimensions go entirely to Si baseline, which leads Novelty, Significance, Effectiveness, Excitement, and Originality. ResearchAgent ranks last on 7 of 10 dimensions, with Feasibility at 4.24 and Reproducibility at 4.04, well below the other three systems and consistent with the surface-level retrieval its entity-store paradigm rewards.

\paragraph{Inter-rater agreement.} Krippendorff's $\alpha$ on interval data follows the same execution-versus-creativity split: moderate on the four execution dimensions ($\alpha$ from 0.56 to 0.62), fair on \emph{Overall} ($\alpha = 0.40$), and fair to very fair on the five creativity dimensions ($\alpha$ from 0.05 to 0.27). \texttt{GoR-SFT}'s wins fall on the dimensions where raters most reliably converge, making the execution-side advantage a clearly discriminated human-judgment signal, while the creativity-side gap to Si baseline sits within the noise floor of expert creativity scoring.

\begin{table}[h]
\caption{\textbf{Human evaluation} on a 5-seed blinded subset of the test set, scored by 5 PhD raters on a 1--10 Likert scale (mean over $n{=}25$ per cell). Best per column in \textbf{bold}.}
\label{tab:human-eval}
\centering
\small
\begin{tabular}{lcccccccccc}
\toprule
\textbf{Method} & \textbf{Nov} & \textbf{Sig} & \textbf{Feas} & \textbf{Clar} & \textbf{Eff} & \textbf{Exc} & \textbf{Snd} & \textbf{Orig} & \textbf{Repro} & \textbf{Overall} \\
\midrule
\texttt{GoR-SFT}\,($\star$)   & 5.88          & 6.72          & \textbf{6.96} & \textbf{7.48} & 6.20          & 6.24          & \textbf{7.00} & 5.92          & \textbf{6.92} & \textbf{6.56} \\
\texttt{Si baseline}          & \textbf{6.24} & \textbf{7.04} & 6.00          & 7.08          & \textbf{6.72} & \textbf{6.68} & 6.24          & \textbf{6.24} & 5.96          & 6.48          \\
\texttt{CoI-Agent}            & 5.72          & 6.84          & 5.64          & 6.24          & 6.40          & 6.00          & 5.72          & 5.68          & 6.00          & 6.00          \\
\texttt{ResearchAgent}        & 5.92          & 6.76          & 4.24          & 4.68          & 5.80          & 5.60          & 4.24          & 5.92          & 4.04          & 4.84          \\
\bottomrule
\end{tabular}
\end{table}

\section{Case studies: full idea proposals}
\label{app:case-study}

This appendix collects two qualitative case studies. The first (Case study \#2 below) is a supplementary example beyond the body case in Section~\ref{sec:exp:case} that illustrates a typical narrow-loss outcome. The second (Table~\ref{tab:case-diff}) provides the full proposals for the body \emph{AbGen} case in Section~\ref{sec:exp:case}.

\paragraph{Case study \#2: \emph{Does Reinforcement Learning Really Incentivize Reasoning?}} A second seed paper from the test set, \emph{Does Reinforcement Learning Really Incentivize Reasoning Capacity in LLMs Beyond the Base Model?}~\citep{yue2025does}, retains a subgraph spanning 12 refs across RL methods (DeepSeek-R1, DeepSeekMath, code-r1, PPO, REINFORCE, the Sutton textbook), recent base models (Qwen2.5-1M, Llama 3, OpenAI o1), and reasoning benchmarks (Solving Quantitative Reasoning, OlympiadBench). The frontier gap visible here is that no retained ref combines per-step process supervision (DeepSeekMath / o1-style) with an RLHF training pipeline that learns from human feedback rather than from a fixed reward.

Reading the four proposals (Table~\ref{tab:case-diff-rl}), \texttt{GoR-SFT} proposes HumanEvalRLHF, an RLHF framework with an RNN-based policy gradient and a novel reward approximation scheme, grounded in the PPO and REINFORCE classics from the subgraph. Si baseline proposes RRC (Reinforcement-Reasoner Circuits), a graph-decomposed reasoning step-reward scheme on GSM-8K and MATH. CoI-Agent proposes a multi-dimensional RL framework extending RLHF for reasoning. ResearchAgent proposes ART-FD, combining self-validated token gradients on high-entropy tokens, dynamic theorem creation, and multi-level feedback loops. The LLM-judge verdict: \texttt{GoR-SFT} wins \textbf{14-6} against Si baseline and \textbf{13-7} against CoI-Agent, but loses \textbf{9-11} to ResearchAgent narrowly, primarily on the Feasibility and Clarity dimensions where RA's more elaborate multi-component framework reads as more polished.

This case complements the AbGen case in Section~\ref{sec:exp:case} by illustrating a typical narrow-loss outcome: \texttt{GoR-SFT}'s grounding in the RL classics (PPO, REINFORCE) yields a coherent but more conservative proposal than RA's multi-component synthesis. The two cases together cover both the decisive-win and narrow-loss ends of \texttt{GoR-SFT}'s distribution against the three published baselines.

\begin{table}[h]
\caption{\textbf{Case study \#2 on the \emph{Does RL Really Incentivize Reasoning?} seed: abridged proposed methods.} Pair-Elo is the LLM-judge tournament verdict on this single seed (5-dimension $\times$ 2-ordering $\times$ 2-point scale, max 20).}
\label{tab:case-diff-rl}
\centering
\footnotesize
\renewcommand{\arraystretch}{1.18}
\begin{tabularx}{\linewidth}{@{}p{1.6cm} *{4}{>{\raggedright\arraybackslash}X}@{}}
\toprule
 & \textbf{\texttt{GoR-SFT} (ours)} & \textbf{Si baseline} & \textbf{CoI-Agent} & \textbf{ResearchAgent} \\
\midrule
\textit{Idea label}
 & HumanEvalRLHF
 & RRC: Reinforcement-Reasoner Circuits
 & Multi-Dimensional RL Framework
 & ART-FD: Adaptive Reasoning via Token Gradients \\
\addlinespace[3pt]
\textit{Proposed method (abridged)}
 & RLHF framework that learns an evaluation strategy for scientific reasoning. An RNN models reasoning steps. A policy gradient with a novel reward approximation scheme reduces memory and improves sample efficiency.
 & Graph-based decomposition of reasoning paths (GSM-8K, MATH). Each reasoning step is a node, and intermediate step rewards are assigned via the graph during RL fine-tuning.
 & Three-axis RL framework extending RLHF for reasoning: process rewards, output diversity, and reward calibration. Each axis adds a separate optimization signal.
 & Three components: (i) self-validated token gradients on high-entropy tokens, (ii) dynamic theorem creation for novel reasoning chains, (iii) multi-level feedback loops across token, step, and outcome levels. \\
\addlinespace[3pt]
\textit{Subgraph nodes referenced}
 & PPO, REINFORCE, DeepSeekMath (RL classics)
 & none (free-floating step-reward design)
 & RLHF chronological chain
 & high-entropy token and theorem refs from broader RL-for-reasoning literature \\
\addlinespace[3pt]
\textit{Pair Elo on this seed (vs.\ \texttt{GoR-SFT}, /\,20)}
 & n/a
 & \textbf{14}\,vs\,6 \,\emph{(GoR-SFT win)}
 & \textbf{13}\,vs\,7 \,\emph{(GoR-SFT win)}
 & 9\,vs\,\textbf{11} \,\emph{(RA win)} \\
\bottomrule
\end{tabularx}
\end{table}

\paragraph{Case study \#1 (body): \emph{AbGen}.} Table~\ref{tab:case-diff} below provides the full text of the four proposals summarized in Section~\ref{sec:exp:case}. Each cell reproduces the system's \emph{Proposed Method} field, lightly trimmed for table fit but preserving the proposal's components and named-reference grounding.

\begin{table*}[h]
\caption{\textbf{Case study on the \emph{AbGen} seed: full proposed methods.} Subgraph nodes referenced are retained refs in Fig.~\ref{fig:case-subgraph} that the proposal cites by name. Pair-Elo is the LLM-judge tournament verdict on this single seed (5-dimension $\times$ 2-ordering $\times$ 2-point scale, max 20).}
\label{tab:case-diff}
\centering
\footnotesize
\renewcommand{\arraystretch}{1.18}
\begin{tabularx}{\linewidth}{@{}p{1.6cm} *{4}{>{\raggedright\arraybackslash}X}@{}}
\toprule
 & \textbf{\texttt{GoR-SFT} (ours)} & \textbf{Si baseline} & \textbf{CoI-Agent} & \textbf{ResearchAgent} \\
\midrule
\textit{Idea label}
 & Iterative Refinement + Planning Framework
 & AGCLO: Ablation-Guided Collaborative LLM Optimization
 & AbGen-XAI+: Explainable, Scalable, Adaptive Ablation Studies
 & FDMME: Multi-Modal Embedding Engine \\
\addlinespace[3pt]
\textit{Proposed method}
 & Two-component framework. (i) Iterative refinement module breaks complex tasks into subtasks via an LLM agent (MARG-style multi-agent decomposition), generating hypotheses and writing code with human-in-the-loop verification. (ii) Planning module makes long-range decisions across the multi-step research process. The two modules compose via shared task state and feedback.
 & Three-stage iterative multi-agent system. (i) Iterative hypothesis generation by an experiment-designer LM proposing testable hypotheses about model components. (ii) Dynamic ablation planner that sequences ablation studies covering feature interactions. (iii) Reviewer agent that integrates results and triggers a refinement loop across designer, planner, and reviewer.
 & Extension of AbGen and ML-Bench combining (i) explainability via XAI on each ablation step, (ii) scalability via batched ablation execution, and (iii) adaptive learning that updates the ablation strategy based on prior-run feedback. The proposal stays within the AbGen-plus-ML-Bench chronological chain.
 & Three-stage multi-modal embedding pipeline. (i) Unified Multi-Modal Embedding Engine integrates text, schematic diagrams, and tables via CLIP / MISS-QA-style models. (ii) Multi-document context aggregator. (iii) Embedding-augmented LLM produces ablation studies. The proposal pivots to a multimodal direction outside the core ablation-design subgraph. \\
\addlinespace[3pt]
\textit{Subgraph nodes referenced}
 & MARG (1 explicit anchor)
 & none (free-floating multi-agent design)
 & AbGen, ML-Bench (chronological chain)
 & MISS-QA, schematic-diagram refs (multimodal subset) \\
\addlinespace[3pt]
\textit{Pair Elo on this seed (vs.\ \texttt{GoR-SFT}, /\,20)}
 & n/a
 & \textbf{15}\,vs\,5 \,\emph{(GoR-SFT win)}
 & \textbf{15}\,vs\,5 \,\emph{(GoR-SFT win)}
 & 10\,vs\,10 \,\emph{(tie)} \\
\bottomrule
\end{tabularx}
\end{table*}


\end{document}